\documentclass{article}

\PassOptionsToPackage{square,numbers,sort&compress}{natbib}

\usepackage[preprint]{neurips_2023}

\usepackage[utf8]{inputenc} 
\usepackage[T1]{fontenc}    
\usepackage{hyperref}       
\usepackage{url}            
\usepackage{booktabs}       
\usepackage{amsfonts}       
\usepackage{nicefrac}       
\usepackage{microtype}      
\usepackage{xcolor}         

\usepackage{amsmath}
\usepackage{amssymb}
\usepackage{amsthm}

\usepackage{algorithm}
\usepackage{algorithmic}
\usepackage{graphicx}
\graphicspath{{./output/}}
\DeclareGraphicsExtensions{.png}
\usepackage{multirow}
\def\Appref#1{Appendix~\ref{#1}}
\def\halfcheck{$\checkmark{}\mkern-13mu{\smallsetminus}$}
\def\notcheck{}

\usepackage{amsmath,amsfonts,bm}

\def\eqref#1{equation~\ref{#1}}

\def\1{\bm{1}}

\def\vtheta{{\bm{\theta}}}

\def\vh{{\bm{h}}}

\def\vx{{\bm{x}}}
\def\vy{{\bm{y}}}
\def\vz{{\bm{z}}}

\def\mA{{\bm{A}}}

\def\mH{{\bm{H}}}
\def\mI{{\bm{I}}}

\def\mM{{\bm{M}}}

\def\mX{{\bm{X}}}
\def\mY{{\bm{Y}}}
\def\mZ{{\bm{Z}}}

\DeclareMathAlphabet{\mathsfit}{\encodingdefault}{\sfdefault}{m}{sl}
\SetMathAlphabet{\mathsfit}{bold}{\encodingdefault}{\sfdefault}{bx}{n}

\def\sD{{\mathbb{D}}}

\def\sV{{\mathbb{V}}}

\newcommand{\E}{\mathbb{E}}
\newcommand{\Ls}{\mathcal{L}}

\title{Graph Entropy Minimization for Semi-supervised Node Classification}

\author{%
  Yi Luo\\
  University of Electronic Science and Technology of China\\
  \texttt{cf020031308@163.com}
  \And
  Guangchun Luo\\
  University of Electronic Science and Technology of China\\
  \And
  Ke Qin\\
  University of Electronic Science and Technology of China\\
  \And
  Aiguo Chen\thanks{Correspongin Author: Aiguo Chen}\\
  University of Electronic Science and Technology of China\\
  \texttt{agchen@uestc.edu.cn}
}

\begin{document}

\maketitle

\begin{abstract}
    Node classifiers are required to comprehensively reduce prediction errors, training resources, and inference latency in the industry.
    However, most graph neural networks (GNN) concentrate only on one or two of them.
    The compromised aspects thus are the shortest boards on the bucket, hindering their practical deployments for industrial-level tasks. 
    This work proposes a novel semi-supervised learning method termed Graph Entropy Minimization (GEM) to resolve the three issues simultaneously.
    GEM benefits its one-hop aggregation from massive uncategorized nodes, making its prediction accuracy comparable to GNNs with two- or more-hops message passing. 
    It can be decomposed to support stochastic training with mini-batches of independent edge samples, achieving extremely fast sampling and space-saving training.
    While its one-hop aggregation is faster in inference than deep GNNs, GEM can be furtherly accelerated to an extreme by deriving a non-hop classifier via online knowledge distillation.
    Thus, GEM can be a handy choice for latency-restricted and error-sensitive services running on resource-constraint hardware.
    Code is available at \url{https://github.com/cf020031308/GEM}.
\end{abstract}

\section{Introduction}

Node classification is a commonly met task in graphic systems involving data points and their relationships.
Its goal is to classify data points (or nodes) by assigning labels to them with some of them already categorized.
Successful real-world applications of node classification include property regression with molecular graphs~\cite{Gilmer2017NeuralMP}, speed forecasting in traffic networks~\cite{DBLP:conf/ijcai/WuPLJZ19}, field matching for submitted manuscripts in citation networks~\cite{DBLP:conf/nips/HuFZDRLCL20}, and image classification~\cite{Liu2019PrototypePN}.
In recent decades, graph neural networks (GNN)~\cite{DBLP:journals/tnn/WuPCLZY21} become conspicuous both in academics and industry due to their dominant performance in node classification.
Their underline principle is message passing.
By repeatedly aggregating and transforming information from local neighbourhoods, GNNs can get informative node representations considering both node features and graph structures.
While many advanced GNNs are continuously proposed and studied in academics, most of them are inevitably undermined in industrial-level tasks when confronting scarce labels, constrained training resources, and restricted inference latency.

The first issue is that labels are scarcely distributed in industrial graphs because collecting them often involves expert knowledge and data privacy~\cite{DBLP:conf/nips/BerthelotCGPOR19}.
By contrast, unlabeled data are far more abundant to access.
Therefore, learning with both precious labels and massive unlabeled data in semi-supervised settings~\cite{DBLP:conf/iclr/KipfW17} is crucial to reduce prediction errors in real-world node classification.
Existing GNNs achieve this by fitting the ground-truth labels using information from their receptive fields covering nearby unlabeled nodes.
However, the range of receptive fields is empirically limited to avoid oversmoothing~\cite{DBLP:conf/iccv/Li0TG19} and overfitting~\cite{DBLP:conf/aaai/DingWC022}, resulting in low utilization of unlabeled data in industrial graphs with scarce labels.
A more practical way to utilize all unlabeled nodes than deepening GNNs is to integrate unsupervised learning objectives~\cite{DBLP:conf/nips/HamiltonYL17}.
For example, GraphMLP~\cite{DBLP:journals/corr/abs-2106-04051} uses a multi-layered perceptron (MLP) to fit ground-truth labels with contrastive objectives~\cite{DBLP:journals/tkde/WuLTGL23}.
It forces the intermediate representations of adjacent nodes to be close and pulls away representations of nodes that are disconnected.
The contrastive objectives regularize representations of nodes in the entire graph, making the predictions robust and general.

The second issue is that graphs in the industry are large, making training GNNs memory-expensive~\cite{DBLP:conf/iclr/JinZZLTS22}.
Stochastic training methods are successively proposed to largely reduce memory footprint by mini-batch training~\cite{DBLP:journals/ieeejas/LiuYDLYF22}.
For example, GraphSAGE~\cite{DBLP:conf/nips/HamiltonYL17}, FastGCN~\cite{DBLP:conf/iclr/ChenMX18}, and GraphSAINT~\cite{DBLP:conf/iclr/ZengZSKP20} move the training from the entire large-scale graph to small subgraphs.
Edge Convolutional Network (ECN, see~\Appref{app:ecn}) points out that their sampling procedures are time-consuming in subgraph formation.
It then proposes edge-wise mini-batch training to avoid subgraph formation, accelerating its sampling to an extrem.

The third issue is that high speed or low latency in inference is crucial for real-world applications~\cite{DBLP:journals/pvldb/ZhouSZKP21}.
Previous studies~\cite{Yan2020TinyGNNLE, DBLP:conf/iclr/ZhangLSS22} show that the inference latency of GNN exponentially grows as the number of layers increases.
Therefore, despite the superior classification performance of deep GNNs, light classifiers with one-~\cite{Cen2019RepresentationLF, Li2019GraphIN} or non-hop~\cite{DBLP:conf/iclr/ZhengH0KWS22} aggregations are still the major workhorses in industrial-level applications.
The performance gap between deep GNNs and light classifiers is recently filled by knowledge distillation~\cite{DBLP:journals/ijcv/GouYMT21}.
Light classifiers are trained in topology-rich contexts inherited from deep GNNs to gain high performance while they still enjoy their fast inference with short-range aggregations.
For example, TinyGNN~\cite{Yan2020TinyGNNLE} distils knowledge into a classifier accessing information from only one-hop neighbourhoods and GLNN~\cite{DBLP:conf/iclr/ZhangLSS22} distils knowledge into a topology-free MLP\@.
Both of them gain performances comparable to deep GNNs while they are inherently fast in inference.

Although a growing number of methods have been proposed to address these issues, each of them only tackles one or two issues but suffers from the others.
For example, GNNs with advanced architectures like JKNet~\cite{DBLP:conf/icml/XuLTSKJ18} and GCNII~\cite{DBLP:conf/icml/ChenWHDL20} can effectively reduce prediction errors, but they are memory-expensive to train and slow to infer.
Stochastic training methods reduce training resources with negligible accuracy loss but leave inference with multi-hops message passing unaccelerated.
Knowledge distillation derives small-scale student models with low inference latency, but the teacher model requires additional resources and the performance is not always guaranteed.
Thus, these methods are far from perfect to deploy in industrial environments that are error-sensitive, resource-constraint, and latency-restricted.

\begin{table}\small
\caption{Properties of node classification methods concerning industrial-level tasks}\label{tbl:methods}
\centering
\begin{tabular}{lccccc}
\toprule
    Method                             & Data       & Mini-batch & subgraph-free & one-hop    & non-hop    \\
                                       & utility    & training   & sampling      & inference  & inference  \\
\midrule                                                                          
    MLP                                & \notcheck  & \checkmark & \checkmark    & \notcheck  & \checkmark \\
    GCN/JKNet/GCNII                    & \halfcheck & \notcheck  & \notcheck     & \notcheck  & \notcheck  \\
    GraphSAGE/FastGCN/GraphSAINT       & \halfcheck & \checkmark & \notcheck     & \notcheck  & \notcheck  \\
    ECN                                & \halfcheck & \checkmark & \checkmark    & \notcheck  & \notcheck  \\
    GLNN                               & \checkmark & \checkmark & \checkmark    & \notcheck  & \checkmark \\
    TinyGNN                            & \checkmark & \notcheck  & \notcheck     & \checkmark & \notcheck  \\
    GraphMLP                           & \checkmark & \checkmark & \notcheck     & \notcheck  & \checkmark \\
    \textbf{GEM} (\ref{sec:gem})       & \checkmark & \notcheck  & \notcheck     & \checkmark & \notcheck  \\
    \textbf{EEM} (\ref{sec:eem})       & \checkmark & \checkmark & \checkmark    & \checkmark & \notcheck  \\
    \textbf{OKDEEM} (\ref{sec:okdeem}) & \checkmark & \checkmark & \checkmark    & \checkmark & \checkmark \\
\bottomrule
\end{tabular}
\end{table}

In this work, we propose a series of methods to resolve all the aforementioned industrial-level issues simultaneously.
Specifically, we propose Graph Entropy Minimization (GEM) to utilize unlabeled nodes.
GEM is effective in semi-supervised node classification where labels are scarce.
It implicitly helps label information to propagate further that GEM with only one-hop aggregation can surpass deep GNNs with two- or more-hop message passing.
Then, we derive the edge-wise mini-batch training method of GEM termed Edge Entropy Minimization (EEM).
Inherited from ECN, EEM is space-saving in training and extremely fast in sampling.
While GEM and EEM are inherently fast in their one-hop inferences, we furtherly accelerate them by integrating Online Knowledge Distillation (OKD) to derive OKDEEM which supports both one- and non-hop inferences.
The properties of baselines and our methods are summarized in Table~\ref{tbl:methods}.

\section{Related works}


\subsection{Entropy minimization}

Entropy Minimization (EM)~\cite{DBLP:conf/nips/GrandvaletB04} is a light-weight method to benefit supervised learning by making the model confident in predicting on unlabeled data~\cite{DBLP:conf/nips/BerthelotCGPOR19}.
Applying it to node classification, EM is to append the following regularization term to the loss of supervised learning:
\[
	\E_{\tilde \sV} \Ls(f_\vtheta(\mA, \mX), \sigma(f_\vtheta(\mX) / \tau)),
\]
where $\tilde \sV$ is a sampled subset of all nodes $\sV$, $\Ls$ is the cross entropy function, $f_\vtheta$ is a GNN for graph $G = (\mA, \mX)$, $\sigma$ is the softmax function, and $\tau$ is a temperature scaler to sharpen predictions~\cite{DBLP:conf/icml/GuoPSW17}.
EM is light-weight in two respects.
First, in the framework of maximum a posteriori estimation (MAP), EM only assumes that the relationship between data points $\mX$ and labels $\mY$ can still holds beyond the training set $\sD$.
This assumption is already made when using $f_\vtheta$ trained on $\sD$ to predict labels for nodes out of $\sD$.
Thus, EM introduces no additional assumptions that it can coorporate any model $f_\vtheta$ to work on graphs with varying data distributions.
Second, EM only enlarges the evaluation scope from the training set for supervised learning to its union with $\tilde \sV$.
The complexity increase is trivial if $\tilde \sV$ is small or the evaluation scope is already large, such as in full-batch training.
Thus, EM can be applied to arbitrary supervised learning applications to leaverage unlabeled data without introducing much burden.

\subsection{Edge-wise stochastic training}

A GNN with its last layer to be a graph convolutional layer~\cite{DBLP:conf/iclr/KipfW17} is formulated as
\begin{equation}\label{eq:gnn}
    \mbox{GNN} (\mA, \mX) = \sigma(\tilde \mA \mH), \quad \mH = f_\vtheta(\mA, \mX),
\end{equation}
where $\tilde \mA$ is the normalized version of adjacency matrix $\mA$, and $f_\vtheta$ is an arbitrary model.

Edge Convolutional Network (ECN, see~\Appref{app:ecn}) demonstrates that the edge-wise loss of GNN is the upper bound of its node-wise loss, as
\[
    \E_{v_i \in \sD} \Ls(\sum\limits_{v_j \in N(v_i)} \tilde A_{ij} \vh_j, \vy_i) \le \E_{(v_i, v_j) \in E, v_i \in \sD} \tilde A_{ij} \Ls(\vh_j, \vy_i)
\]
where $N(v_i)$ is the neighbourhood of node $v_i$, $\tilde A_{ij}$ is the element of $\tilde \mA$ at its $i$-th row and $j$-th column, $E$ is the edge set, $\vh_j$ is the $j$-th row of $\mH$, and $\vy_i$ is the $i$-th row of $\mY$.
Minimizing the edge-wise loss can also optimize GCN with bounded non-linearity differences.

It then proposes to train GCN with mini-batches of edges.
Like in conventional stochastic training methods, the edge batch can be small to reduce memory footprint in training.
Most importantly, the edges can be independently sampled without the need to form a subgraph.
The elimination of subgraph formation in ECN reduces time consumption in sampling by orders of magnitude without compromising prediction accuracy and convergence speed.

\subsection{Online knowledge distillation}

Knowledge distillation~\cite{DBLP:journals/ijcv/GouYMT21} is to transfer the knowledge from a well-trained, parameter-freeze teacher model to a small-scaled, optimizable student model by aligning their outputs or intermediate representations.
After knowledge distillation, the student model can have inference performance comparable to the teacher model.
Due to their small scales and simple structures, the student models are usually deployed to industrial platforms to infer at speed.
Recent studies utilize knowledge distillation to derive shallow GNNs~\cite{Yan2020TinyGNNLE} or even MLPs~\cite{DBLP:conf/iclr/ZhangLSS22} that inherit performance from deep GNNs for deployment purposes.
However, a large-capacity teacher model may not always be available.
Online knowledge distillation is thus proposed to update one or multiple student models without a teacher model.
An effective way to achieve this is to induce predictions that are consistent with relevant data such as the universal label distribution of same-labelled nodes~\cite{DBLP:conf/cvpr/YunPLS20}.

\section{Methodology}

In this section, we propose Graph Entropy Minimization (GEM) to alleviate label insufficiency, Edge Entropy Minimization (EEM) to train GEM in mini-batch, and Online Knowledge Distillation EEM (OKDEEM) to derive a non-hop classifier.

\subsection{Utilizing unlabeled data with graph entropy minimization (GEM)}\label{sec:gem}

\begin{algorithm}[!t]
    \caption{Graph entropy minimization}\label{algo:gem}
    \begin{algorithmic}[1]
        \REQUIRE GNN or MLP $f_\vtheta$, adjacency matrix $\mA$, feature matrix $\mX$, training set indicator $\mM_t$, ground-truth labels $\mY$, hyperparameter $\lambda$, temperature scaler $\tau$.
        \STATE Compute logits for predictions $\mZ = \tilde \mA \mH = \mA f_\vtheta(\mA, \mX)$
        \STATE Compute pseudo labels $\tilde \mY = (\mI - \mM_t) \sigma(\mZ / \tau) + \mM_t \mY$
        \STATE Minimize $\Ls(\mM_t \mZ, \mM_t \mY) + \lambda \Ls(\mH, \tilde \mA^T \tilde \mY)$
    \end{algorithmic}
\end{algorithm}

According to the literature of Edge Convolutional Network (ECN, see~\Appref{app:ecn}), the regularization term of entropy minimization (EM) in GNN defined as Equation~\ref{eq:gnn} has an upper bound as
\begin{equation}\label{eq:gem}
    \Ls(\tilde \mA \mH, \tilde \mY) \le \Ls(\mH, \tilde \mA^T \tilde \mY), \quad \tilde \mY = (\mI - \mM_t) \hat \mY + \mM_t \mY, \quad \hat \mY = \sigma(\tilde \mA \mH / \tau),
\end{equation}
where $\mM_t$ is an 0-1 diagonal matrix indicating the training set.
Thus, minimizing the right part $\Ls(\mH, \tilde \mA^T \tilde \mY)$ of the inequality also reduces the EM term, benefiting the supervised learning of $f_\vtheta$ from all unlabeled data without introducing additional assumption and complexity.
Supervised graph learning regularized by $\Ls(\mH, \tilde \mA^T \tilde \mY)$ is termed as Graph Entropy Minimization (GEM) in this work.
Algorithm~\ref{algo:gem} describes its training epoch.

We demonstrate that GEM is to search for an equilibrium $\tilde \mY$ to meet
\begin{equation}\label{eq:geq}
	\tilde \mY = (\mI - \mM_t) \sigma( \tilde \mA \log \sigma( \tilde \mA^T \tilde \mY ) / \tau ) + \mM_t \mY.
\end{equation}
This equation implies that the predictions on unlabeled nodes are potentially influenced by pseudo labels from their two-hop neighbourhoods. 
Thus, ground-truth labels $\mM_t \mY$ are implicitly propagated two- or more-hop away under the effect of prediction equilibrium~\cite{DBLP:journals/corr/abs-2211-10629}.
Therefore, GEM helps message passing and alleviates label insufficiency.

\subsection{Training GEM with fast-sampled edges}\label{sec:eem}

Decomposing the optimization objective in Algorithm~\ref{algo:gem} into edges (see \Appref{app:ecn}), we get the edge-wise loss of GEM as
\begin{equation}\label{eq:eem}
    \E_{p(v_i, v_j) \propto \tilde A_{ij}, v_i \in \sD} \Ls(\vh_j, \vy_i) + \lambda \E_{p(v_i, v_j) \propto \tilde A_{ij}} \Ls(\vh_j, \tilde \vy_i).
\end{equation}
Thus, when the network $f_\vtheta$ is an MLP, we can optimize GEM by minimizing Equation (\ref{eq:eem}) in mini-batches of edges like ECN.

The only question is that $\tilde \vy_i$ is not available in the edge-wise stochastic training where the neighbourhood $N(v_i)$ of a specific node $v_i$ is not guaranteed to be completely sampled.
To resolve this issue, we estimate the logits $\vz_i = \sum\limits_{v_j \in N(v_i)} \tilde A_{ij} \vh_j$ by accumulating $\vh_j$ in training and adopt exponential moving average (EMA) of $\vz_i$ to approximate $\hat \vy_i = \sigma(\sum\limits_{v_j \in N(v_i)} \tilde A_{ij} \vh_j / \tau) \approx \sigma(\vz_i + (1 - \tau) \vz_i')$, where $\vz_i'$ is the historical value of $\vz_i$ in the last epoch of training.

\begin{algorithm}[!t]
    \caption{Edge entropy minimization}\label{algo:eem}
    \begin{algorithmic}[1]
        \REQUIRE MLP $f_\vtheta$, adjacency matrix $\mA$, feature matrix $\mX$, training set indicator $\mM_t$, ground-truth labels $\mY$, hyperparameter $\lambda$, temperature scaler $\tau$.
        \STATE Prepare EMA $\bar \mZ = \log(\mM_t \mY)$
        \STATE Compute the average degree $d = | E | / | \sV |$
        \REPEAT[training epoch]
        \REPEAT[mini-batch training]
        \STATE Get node indicators $\mM_i$ and $\mM_j$ by edge sampling of $p(v_i, v_j) \propto \tilde A_{ij}$
        \STATE Compute partial peer logits $\mH_i = f_\vtheta(\mM_i \mX)$
        \STATE Minimize $\Ls(\mM_t \mH_i, \mM_t \mM_j \mY) + \lambda \Ls(\mH_i, \sigma(\mM_j \bar \mZ))$
        \STATE Update EMA $\mM_j \bar \mZ := \mM_j \bar \mZ + \mH_i / d$
        \UNTIL{$|E|$ edges are sampled in total}
        \STATE Update EMA $\bar \mZ := \bar \mZ \cdot (1 - \tau)$
        \UNTIL{training stops}
    \end{algorithmic}
\end{algorithm}

Algorithm~\ref{algo:eem} describes the edge-wise training phase of GEM.
In line 1, we initialize EMA $\bar \mZ$ with the logits of ground-truth labels $\mM_t \mY$.
In line 5, we cast edge sampling to get independent edge samples $(v_i, v_j) \in E$ with $\mM_i$ indicating nodes $v_i$ and $\mM_j$ indicating their peers $v_j$.
In line 6, a part of peer logits $\vh_i$ for $\hat \vy_j$ is computed using features $\vx_i$ of node $v_i$.
For simplicity, only features of nodes indicating by $\mM_i$ is considered in this description.
In line 7, $\vh_i$ is supervised by $\vy_j$ if node $v_j$ is from the training set.
Meanwhile, $\vh_i$ is regularized by the EMA-estimated pseudo labels $\sigma(\bar \vz_j)$.
In line 8, we update EMA by adding $\mH_i$ to it with a scale of $1/d$ because each node is expected to be sampled $d$ times with a sampling budget of $|E|$ due to importance sampling. 
In line 10, we scale EMA by $1 - \tau$.
This is equivalent to shapening pseudo labels with $\tau$ in node-wise training.

The algorithm is named Edge Entropy Minimization (EEM).
Inherited from ECN, EEM is space-saving in training and extremely fast in sampling.

\subsection{Inference with one- and non-hop aggregations}\label{sec:okdeem}

We propose another option to estimate $\hat \mY$ in EEM based on Online Knowledge Distillation (OKD).
Given another MLP $g_\varphi$ to fit $\hat \mY = \sigma(\tilde \mA \mH / \tau)$, its optimization objective is to minimize the difference between $\log \sigma(g_\varphi(\vx_i)), \forall v_i \in \sV$ and the mean of $\log \sigma(\vh_j / \tau), v_j \in N(v_i)$.
Thus, in edge-wise training, we minimize the difference between $\log \sigma(g_\varphi(\vx_i))$ and $\log \sigma(\vh_j / \tau)$ for every sampled edge $(v_i, v_j)$ to complete online knowledge distillation. 

\begin{algorithm}[!t]
    \caption{Online knowledge distillation and edge entropy minimization}\label{algo:okdeem}
    \begin{algorithmic}[1]
        \REQUIRE MLP $f_\vtheta$, adjacency matrix $\mA$, feature matrix $\mX$, training set indicator $\mM_t$, ground-truth labels $\mY$, hyperparameter $\lambda, \alpha$, temperature scaler $\tau$.
        \STATE Get node indicators $\mM_i$ and $\mM_j$ by edge sampling of $p(v_i, v_j) \propto \tilde A_{ij}$
        \STATE Compute partial peer logits and target logits \[(\mH_i, \hat \mZ_i) = f_\vtheta(\mM_i \mX), \quad (\mH_j, \hat \mZ_j) = f_\vtheta(\mM_j \mX)\]
        \STATE Compute pseudo labels \[\begin{cases}
            \tilde \mY_j^{(i)} = (\mI - \mM_t \mM_j) \sigma(\mH_i / \tau) + \mM_t \mM_j \mY \\
            \tilde \mY_i^{(i)} = (\mI - \mM_t \mM_i) \sigma(\hat \mZ_i / \tau) + \mM_t \mM_i \mY \\
            \tilde \mY_i^{(j)} = (\mI - \mM_t \mM_i) \sigma(\mH_j / \tau) + \mM_t \mM_i \mY \\
            \tilde \mY_j^{(j)} = (\mI - \mM_t \mM_j) \sigma(\hat \mZ_j / \tau) + \mM_t \mM_j \mY \\
        \end{cases}\]
        \STATE Compute the supervising loss \[L_{SL} = \Ls(\mM_t (\mH_i + \hat \mZ_j), \mM_t \mM_j \mY) + \Ls(\mM_t (\mH_j + \hat \mZ_i), \mM_t \mM_i \mY)\]
        \STATE Compute the entropy minimization loss $L_{EM} = \Ls(\mH_i, \tilde \mY_j^{(j)}) + \Ls(\mH_j, \tilde \mY_i^{(i)})$
        \STATE Compute the knowledge distillation loss $L_{KD} = \Ls(\hat \mZ_i, \tilde \mY_i^{(j)}) + \Ls(\hat \mZ_j, \tilde \mY_j^{(i)})$
        \STATE Minimize $L_{SL} + \lambda L_{EM} + \alpha L_{KD}$
    \end{algorithmic}
\end{algorithm}

Emperically, $f_\vtheta$ and $g_\varphi$ can be combined into a unified MLP which outputs $\begin{bmatrix}\mH & \log \sigma(\hat \mZ)\end{bmatrix}$.
This derives OKDEEM described by Algorithm~\ref{algo:okdeem}.
In line 1, we use an MLP to output partial peer logits $\mH_i, \mH_j$ and target logits $\hat \mZ_i, \hat \mZ_j$.
$\mH_i$ and $\hat \mZ_i$ are computed using features of nodes $v_i$ indicated by $\mM_i$, predicting labels of peer nodes $v_j$ and of themselves, respectively.
Likewise, $\mH_j$ and $\hat \mZ_j$ are from node $v_j$, predicting labels of node $v_i$ and $v_j$, respectively.
In line 3, the aforementioned label predictions are computed and overwritten by ground-truth labels.
In line 4, ground-truth labels are supervising outputs of $f_\vtheta$ according to what they are predicting.
In line 5, entropy minimization term is computed like in EEM.
In line 6, online knowledge distillation is adopted in the edge-wise manner.

After training, we get an MLP that can be deployed to industrial platforms to cast both one- and non-hop inferences.
For example, in citation networks, a submitted manuscript $v_i$ is categoried by its content $x_i$ and its references $v_j \in N(v_i)$.
OKDEEM can complete this one-hop inference using $\frac{1}{|N(v_i) + 1|} \sum\limits_{v_j \in N(v_i) \cup \{v_i\}} f(x_j)$, where $f$ outputs the first half of the results from $f_\vtheta$.
In social networks, a newly joined user $v_i$ is tagged by its properties $x_i$ without knowing its related people.
OKDEEM can infer his tags with $f_0(x_i)$ where $f_0$ outputs the second half of the results from $f_\vtheta$.

\section{Experiments}\label{sec:experiments}

In this section, we verify the effectiveness of GEM in node classification with scarce labels and reveal its reason: implicitly helping label propagation.
Then, we discuss when GEM will fail to work.

\subsection{One-hop GEM and multi-hop message passing}

\begin{figure*}[t]
\centering
\includegraphics[width=1.0\linewidth]{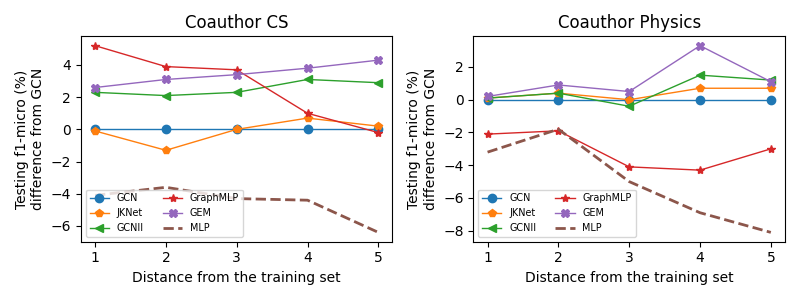}
\caption{
    Averaged testing f1-micro scores (\%) compared with GCN in 10 runs of node classification.
    Nodes for testing are divided by their distances from the training set.
}\label{fig:hop}
\end{figure*}

This experiment verifies whether GEM can propagate label information beyond its one-hop receptive field, as Section~\ref{sec:gem} claims.
For a given graph, we construct the training set by sampling 20 nodes from each class.
Then, we compute the distances between the rest nodes and the training set, dividing them into different groups by distance.
100 nodes are sampled from each group to consist of the testing set.
Finally, 500 random nodes are selected to form the validation set.
We evaluate MLP, GCN, JKNet, GCNII, GraphMLP, and GEM on real-world datasets Coauthor CS and Coauthor Physics~\cite{DBLP:journals/qss/WangSHWDK20}.
The results are depicted in Figure~\ref{fig:hop}, in which a data point represents the difference between the accuracy of the evaluated method and that of GCN.
As the figure shows, there are clear upward trends of curves for GEM and deep GNNs including JKNet and GCNII, meaning that they are more resistant to the performance loss caused by increasing distance than GCN.
In other words, when classifying a node, GEM and deep GNNs can leverage supervising information from a further distance than GCN.  
While this is obvious for deep GNNs, GEM achieves a superior effect using only one-hop aggregation because its regularization term implicitly helps label propagation, as Equation~\ref{eq:geq} suggests.

\subsection{Semi-supervised inductive node classification with scarce labels}\label{sec:node}

This experiment evaluates the performance of different methods on node classification tasks with realistic settings.
Datasets include Cora, Citeseer, Pubmed~\cite{DBLP:journals/aim/SenNBGGE08}, Amazon Photo, Amazon Computer~\cite{DBLP:conf/sigir/McAuleyTSH15}, Coauthor CS, Coauthor Physics, and OGB-Arxiv~\cite{DBLP:conf/nips/HuFZDRLCL20}.
These datasets are different in many aspects, including scale and density.
For Cora, Citeseer, and Pubmed, we adopt their public data splits and transductive settings.
For other datasets, we split them for training, validating, and testing following the semi-supervised settings of Cora, Citeseer, and Pubmed.
In detail, 20 nodes from each class are sampled for training.
500 and 1000 random nodes form the validation and testing set.
The semi-supervised settings simulate the real-world situation where the majority of nodes are unlabeled.
Edges associated with nodes in the validation and testing sets are masked during training following the inductive settings~\cite{DBLP:conf/iclr/ChenMX18} because unseen nodes are commonly met in real-world scenarios.

\begin{table}\small
\caption{
    Accuracy scores (\%) of node classification with real-world datasets.
    Each score is averaged in 10 runs with identical experiment settings.
    Standard deviations are reported in \Appref{app:stderr}.
    Methods are categorized into three groups: deep GNNs, stochastic GNNs, and MLP-based methods.
    The highest score in each group for each dataset is bolded.
}\label{tbl:acc}
\centering
\begin{tabular}{lcccccccc}
\toprule
 & \textbf{Cora} & \textbf{Citeseer} & \textbf{Pubmed} & \textbf{Amazon} & \textbf{Amazon} & \textbf{Coauthor} & \textbf{Coauthor} & \textbf{OGB-} \\
 & & & & \textbf{Photo} & \textbf{Computer} & \textbf{CS} & \textbf{Physics} & \textbf{Arxiv} \\

\textbf{\#Nodes}    & 2708      & 3327      & 19717     & 7650       & 13752      & 18333      & 34493      & 169343 \\
\textbf{\#Edges}    & 5278      & 4552      & 44324     & 119081     & 245861     & 81894      & 247962     & 1166243 \\
\textbf{\#Features} & 1433      & 3703      & 500       & 745        & 767        & 6805       & 8415       & 128 \\
\textbf{\#Classes}  & 7         & 6         & 3         & 8          & 10         & 15         & 5          & 40 \\
\textbf{Label Rate} & 5.17\%    & 3.61\%    & 0.30\%    & 2.09\%     & 1.45\%     & 1.64\%     & 0.29\%     & 0.47\% \\
\textbf{Inductive}  & \notcheck & \notcheck & \notcheck & \checkmark & \checkmark & \checkmark & \checkmark & \checkmark \\
\midrule
GCN      & 81.81   & 69.51   & 76.44   & 91.15   & 80.97   & 89.66   & 92.93   & 53.51   \\
JKNet    & 81.83   & 68.79   & 76.68   & 91.09   & 81.47   & 89.91   & 92.87   & 52.30   \\
GCNII    & 80.50   & 67.09  & 77.24   &\textbf{92.41}  & 82.23   &\textbf{92.70}  & 93.81   & 54.10   \\
GEM      &\textbf{83.05}  &\textbf{74.20}  &\textbf{78.48}  & 92.37   &\textbf{83.69}  & 92.58   &\textbf{94.44}  &\textbf{56.86}  \\
\midrule
GraphSAGE & 78.72  & 68.28   & 75.04  & 91.20   & 79.35   & 89.18  & OOM    & 48.65   \\
ECN      & 80.25   & 67.45  & 75.82  & 90.73   & 80.84   & 88.94  & 91.86   &\textbf{53.90}  \\
EEM      & 81.57   & 72.63   &\textbf{77.49}  & 90.20   &\textbf{82.03}  & 91.44   & 94.01   & 49.78   \\
OKDEEM   &\textbf{82.94}  &\textbf{73.53}  & 76.15   &\textbf{91.71}  & 81.62   &\textbf{91.81}  &\textbf{94.46}  & 51.27   \\
\midrule
MLP      & 49.19  & 45.21  & 67.56  & 72.82  & 60.74  & 85.05  & 87.79  & 30.40  \\
GLNN     & 79.22  & 69.10   & 78.63   &\textbf{88.76}  & 77.99  & 92.27   & |      &\textbf{43.67}  \\
GraphMLP & 79.27   & 72.87   &\textbf{79.91}  & 88.22  & 74.10  &\textbf{92.29}  & 89.32   & OOM    \\
OKDEEM0  &\textbf{82.38}  &\textbf{73.37}  & 76.14   & 88.26  &\textbf{78.31}  & 92.13   &\textbf{94.11}  & 40.83  \\
\bottomrule
\end{tabular}
\end{table}

Evaluated methods include GCN, JKNet, GCNII, GraphSAGE, ECN, MLP, GLNN, GraphMLP, GEM, EEM, and the two inference modes of OKDEEM.
We use OKDEEM to represent its one-hop inference and OKDEEM0 to represent its non-hop inference, described in Section~\ref{sec:okdeem}.
These methods can be categorized into three groups corresponding to the three industrial-level issues this work concerns.
GCN, JKNet, and GCNII are grouped into deep GNNs, which rely on more-hop message passing to utilize more unlabeled data for better predictions.
GraphSAGE, ECN, EEM, and OKDEEM are in the stochastic GNNs group, which reduces training resources with sampling methods.
MLP, GLNN, GraphMLP and OKDEEEM0 are in the MLP-based methods group.
They are extremely fast in inference by ignoring topology information after their deployments.
Every model runs 10 times on a dataset to collect the testing f1-micro score with the highest validation score in each run.
Details about their tuning are reported in \Appref{app:exp}.
The averaged testing f1-micro score is reported in Table~\ref{tbl:acc}.

As we can see from the table, GEM and its derivations (EEM and OKDEEM) match or exceed state-of-the-art baselines in all three groups on most datasets.
The results verify that GEM has advantages in utilizing scarce label information and massive unlabeled data, which are also inherited by EEM and OKDEEM.
Besides, while there are noticeable gaps between OKDEEM and OKDEEM0 in inductive learning tasks, we find the non-hop classifier OKDEEM0 behaves almost the same as the one-hop OKDEEM in transductive settings.
This is because the information of adjacent nodes is explicitly aggregated in the edge-wise training of OKDEEM and transferred to OKDEEM0 via online knowledge distillation.

\subsection{On the speed of EEM and OKDEEM}

\begin{figure*}[t]
\centering
\includegraphics[width=1.0\linewidth]{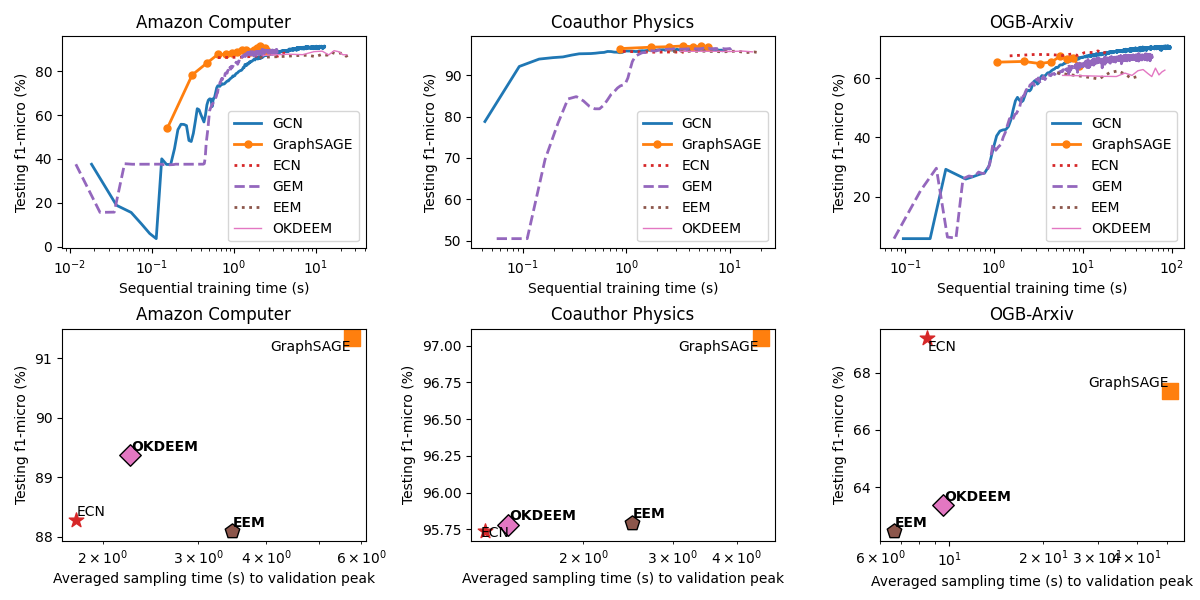}
\caption{
    Time of training and sampling to accuracy peak
}\label{fig:speed}
\end{figure*}

This experiment shows the speed of EEM and OKDEEM in sampling and training.
Figure~\ref{fig:speed} illustrates the sampling and training time for different methods to reach their validation f1-micro peaks.
As is shown, except for GraphSAGE on Amazon Computer, all stochastic training methods, including GraphSAGE, ECN, EEM, and OKDEEM, can meet their performance convergence within only one epoch of training, because these methods are optimized multiple times within one epoch due to mini-batch training.
For sampling, EEM and OKDEEM are slower than ECN in most cases, because EEM and OKDEEM have larger sampling budgets in an epoch to sample from all edges while ECN only samples from edges involving labeled nodes.

\subsection{Inductive node classification with dense labels}

This experiment answers whether the advantage of GEM in utilizing unlabeled data can be offset by dense label distributions, where deep GNNs are easier to utilize all nodes by supervising labels of nodes in the training set.

\begin{figure*}[t]
\centering
\includegraphics[width=1.0\linewidth]{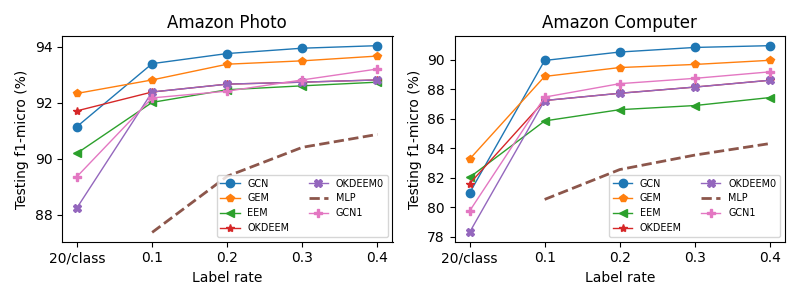}
\caption{
    Averaged testing f1-micro scores (\%) in 10 runs of node classification.
}\label{fig:labels}
\end{figure*}

For a given graph, we construct the training set by sampling a specific per cent of nodes.
Then, the left nodes are divided into the same-sized validation set and testing set.
We evaluate GCN, GEM, EEM, OKDEEM, OKDEEM0, and GCN with one-hop aggregation (GCN1) on Amazon Photo and Amazon Computer.
The results are depicted in Figure~\ref{fig:labels}.
As the figure shows, as the number of ground-truth labels increases in a graph, GEM loses its advantage in utilizing unlabeled data.
However, it is better than the GCN with one-hop aggregation because GEM's regularization term still takes effect.

\section{Conclusions}

This work proposes a series of methods including Graph Entropy Minimization (GEM), Edge Entropy Minimization (EEM), and Online Knowledge Distillation EEM (OKDEEM) to reduce prediction errors, training resources, and inference latency in the industry.
Theory analysis and experiments prove that these methods are effective in utilizing both scarce labels and unlabeled data to reduce prediction errors.
Meanwhile, combined with advanced technologies of edge-wise stochastic training and online knowledge distillation, OKDEEM reduces memory occupation for training, time for sampling, and time for inference.

\begin{ack}
Use unnumbered first level headings for the acknowledgments. All acknowledgments
go at the end of the paper before the list of references. Moreover, you are required to declare
funding (financial activities supporting the submitted work) and competing interests (related financial activities outside the submitted work).
More information about this disclosure can be found at: \url{https://neurips.cc/Conferences/2023/PaperInformation/FundingDisclosure}.

Do {\bf not} include this section in the anonymized submission, only in the final paper. You can use the \texttt{ack} environment provided in the style file to autmoatically hide this section in the anonymized submission.
\end{ack}

\appendix

\section{Edge convolutional networks}\label{app:ecn}

Edge Convolutional Network (ECN) is a work under review.
We supply the manuscript of ECN in supplementary materials.


\section{Experiment details}\label{app:exp}

The real-world datasets in this work are retrieved by PyG~\cite{Fey/Lenssen/2019}, DGL~\cite{wang2019dgl}, and OGB~\cite{DBLP:conf/nips/HuFZDRLCL20}.
In detail, Cora, Citeseer, Pubmed, and OGB-Arxiv are networks of typed articles with citations represented as graph edges.
Amazon Photo and Amazon Computer are segments of the Amazon co-purchase graph~\cite{DBLP:conf/sigir/McAuleyTSH15}, constructed with on-sale goods and their co-purchase relationship.
Coauthor CS and Coauthor Physics are graphs with authors labelled by study fields and connected by co-authorship, based on the Microsoft Academic Graph (MAG)~\cite{DBLP:journals/qss/WangSHWDK20}.

\begin{table}\small
\caption{Hyperparameters for grid searching}\label{tbl:tune}
\centering
	\begin{tabular}{lc}
		\toprule
        Method & Hyperparameters and the searching scope \\
		\midrule
        JKNet & \#layers: $\{ 2, 4, 8 \}$ \\
        GCNII & (\#layers, $\alpha$, $\theta$): $\{2, 4, 8\} \times \{0.1, 0.3, 0.5\} \times \{0.5, 1.0, 1.5\}$ \\
        GLNN & $\alpha$: $\{ 0.1, 1.0 \}$ \\
        GraphMLP & (weight decay, $\tau$, power of $\tilde \mA$, $\alpha$, batch size): \\
        & $\{ 0, 0.005, 0.0005 \} \times \{ 0.5, 1.0, 2.0 \} \times \{ 2, 3 \} \times \{ 1, 10 \} \times \{ 1024, 2048 \}$ \\
        GEM/EEM & (weight decay, \#layers, $\tau$, $\lambda$): \\
        & $\{ 0, 0.005, 0.0005 \} \times \{ 2, 3 \} \times \{ 0.25, 0.5, 0.75 \} \times \{ 0.1, 0.3, 0.5 \}$ \\
        OKDEEM & (weight decay, \#layers, $\tau$, $\lambda$, $\alpha$): \\
        & $\{ 0, 0.005, 0.0005 \} \times \{ 2, 3 \} \times \{ 0.25, 0.5, 0.75 \} \times \{ 0.1, 0.3, 0.5 \} \times \{0.1, 1.0\}$ \\
		\bottomrule
	\end{tabular}
\end{table}

Baselines are imported from PyG or implemented using PyTorch~\cite{NEURIPS2019_9015}.
The dimension of hidden representations for all methods is 256 with a dropout rate 0.5.
Other hyperparameters are grid searched, as summerized in Table~\ref{tbl:tune}.
In training, the Adam optimizer~\cite{DBLP:journals/corr/KingmaB14} is adopted with the learning rate of 0.01.
The early stopping strategy is applied to terminate the training if a model fails to gain performance increase during 100 epochs or iterations.

\section{Standard deviations of experiments reported in Table~\ref{tbl:acc}}\label{app:stderr}.

\begin{table}\small
\caption{
    Standard deviations for accuracy scores (\%) of node classification with real-world datasets.
    Methods are categorized into three groups: deep GNNs, stochastic GNNs, and MLP-based methods.
}\label{tbl:dev}
\centering
\begin{tabular}{lcccccccc}
\toprule
 & Cora & Citeseer & Pubmed & Amazon & Amazon   & Coauthor & Coauthor & OGB- \\
 &      &          &        & Photo  & Computer & CS       & Physics  & Arxiv \\
\midrule
GCN      & 0.38 & 0.56 & 0.25 & 0.98 & 1.96 & 1.58 & 0.81 & 1.90   \\
JKNet    & 0.44 & 0.91 & 0.63 & 0.73 & 1.38 & 1.21 & 1.61 & 1.58   \\
GCNII    & 1.26 & 2.10 & 0.59 & 1.24 & 2.12 & 1.24 & 0.82 & 2.13   \\
GEM      & 0.69 & 0.82 & 0.47 & 0.73 & 1.62 & 0.77 & 0.88 & 1.65   \\
\midrule
GraphSAGE     & 0.87 & 1.18 & 0.58 & 1.16 & 2.52 & 1.46 & OOM  & 1.89   \\
ECN      & 0.87 & 1.08 & 0.79 & 1.27 & 1.93 & 1.42 & 1.05 & 1.51   \\
EEM      & 1.37 & 0.67 & 0.96 & 1.62 & 3.01 & 0.83 & 0.79 & 3.57   \\
OKDEEM   & 1.16 & 1.07 & 0.88 & 1.76 & 2.05 & 0.88 & 0.82 & 1.77   \\
\midrule
MLP      & 1.48 & 1.37 & 1.95 & 2.90 & 3.12 & 2.51 & 2.16 & 2.01  \\
GLNN     & 0.59 & 0.72 & 0.43 & 1.57 & 2.38 & 1.12 & |    & 1.51   \\
GraphMLP & 1.14 & 0.86 & 1.24 & 1.74 & 1.20 & 1.01 & 2.25 & OOM   \\
OKDEEM0  & 0.75 & 1.19 & 1.59 & 2.45 & 2.80 & 0.95 & 0.90 & 1.78  \\
\bottomrule
\end{tabular}
\end{table}

We report standard deviations that are corresponding to Table~\ref{tbl:acc} of Section~\ref{sec:node}.

{
    \small
    \bibliographystyle{apalike}
    \bibliography{linkdist.bib}
}

\end{document}